\def\year{2020}\relax
\documentclass[letterpaper]{article} 
\usepackage{aaai20}  
\usepackage{times}  
\usepackage{helvet} 
\usepackage{courier}  
\usepackage[hyphens]{url}  
\usepackage{graphicx} 
\usepackage{graphicx}  
\usepackage{multirow}
\usepackage{amsmath}
\usepackage[ruled]{algorithm2e} 
\usepackage{amsfonts}
\usepackage{color}
\usepackage{array}
\usepackage[colorlinks = False]{hyperref}

\usepackage{booktabs}

\newcommand{\eg}{\textit{e.g.}, }
\newcommand{\ie}{\textit{i.e.}, }
\newcommand{\etc}{\textit{etc.} }

\title{Pose-Assisted Multi-Camera Collaboration for Active Object Tracking}
\author{
Jing Li*\textsuperscript{\rm 1, 3}, \Large \textbf{Jing Xu*}\textsuperscript{\rm 1 }, \Large \textbf{Fangwei Zhong*}\textsuperscript{\rm 2}, \Large \textbf{Xiangyu Kong}\textsuperscript{\rm 2}, \Large \textbf{Yu Qiao}\textsuperscript{\rm 4}, \Large \textbf{Yizhou Wang}\textsuperscript{\rm 2, 5, 6}\\ 
\textsuperscript{1} Center for Data Science, Peking University 
\\
\textsuperscript{2} Computer Science Dept., Sch’l of EECS, Peking University\\
\textsuperscript{3} Advanced Innovation Center for Future Visual Entertainment(AICFVE), Beijing Film Academy\\
\textsuperscript{4} Key Lab. of System Control and Information Processing (MoE), Shanghai; Automation Dept., Shanghai Jiao Tong University\\
\textsuperscript{5} Center on Frontiers of Computing Studies, Peking University\\
\textsuperscript{6} Deepwise AI Lab\\
\textsuperscript{*} indicates equal contribution\\
lijingg@pku.edu.cn, jing.xu@pku.edu.cn, zfw@pku.edu.cn, kong@pku.edu.cn, qiaoyu@sjtu.edu.cn, yizhou.wang@pku.edu.cn
}

\begin{document}
\maketitle
\begin{abstract}
Active Object Tracking (AOT) is crucial to many vision-based applications, \eg mobile robot, intelligent surveillance. However, there are a number of challenges when deploying active tracking in complex scenarios, \eg target is frequently occluded by obstacles.
In this paper, we extend the single-camera AOT to a multi-camera setting, where cameras tracking a target in a collaborative fashion.
To achieve effective collaboration among cameras, we propose a novel Pose-Assisted Multi-Camera Collaboration System, which enables a camera to cooperate with the others by sharing camera poses for active object tracking.
In the system, each camera is equipped with two controllers and a switcher:
The vision-based controller tracks targets based on observed images. The pose-based controller moves the camera in accordance to the poses of the other cameras. At each step, the switcher decides which action to take from the two controllers according to the visibility of the target.
The experimental results demonstrate that our system outperforms all the baselines and is capable of generalizing to unseen environments.
The code and demo videos are available on our website\sf{ https://sites.google.com/view/pose-assisted-collaboration}.
\end{abstract}

\section{Introduction}
\begin{figure}[tb]
\includegraphics[width=.95\linewidth]{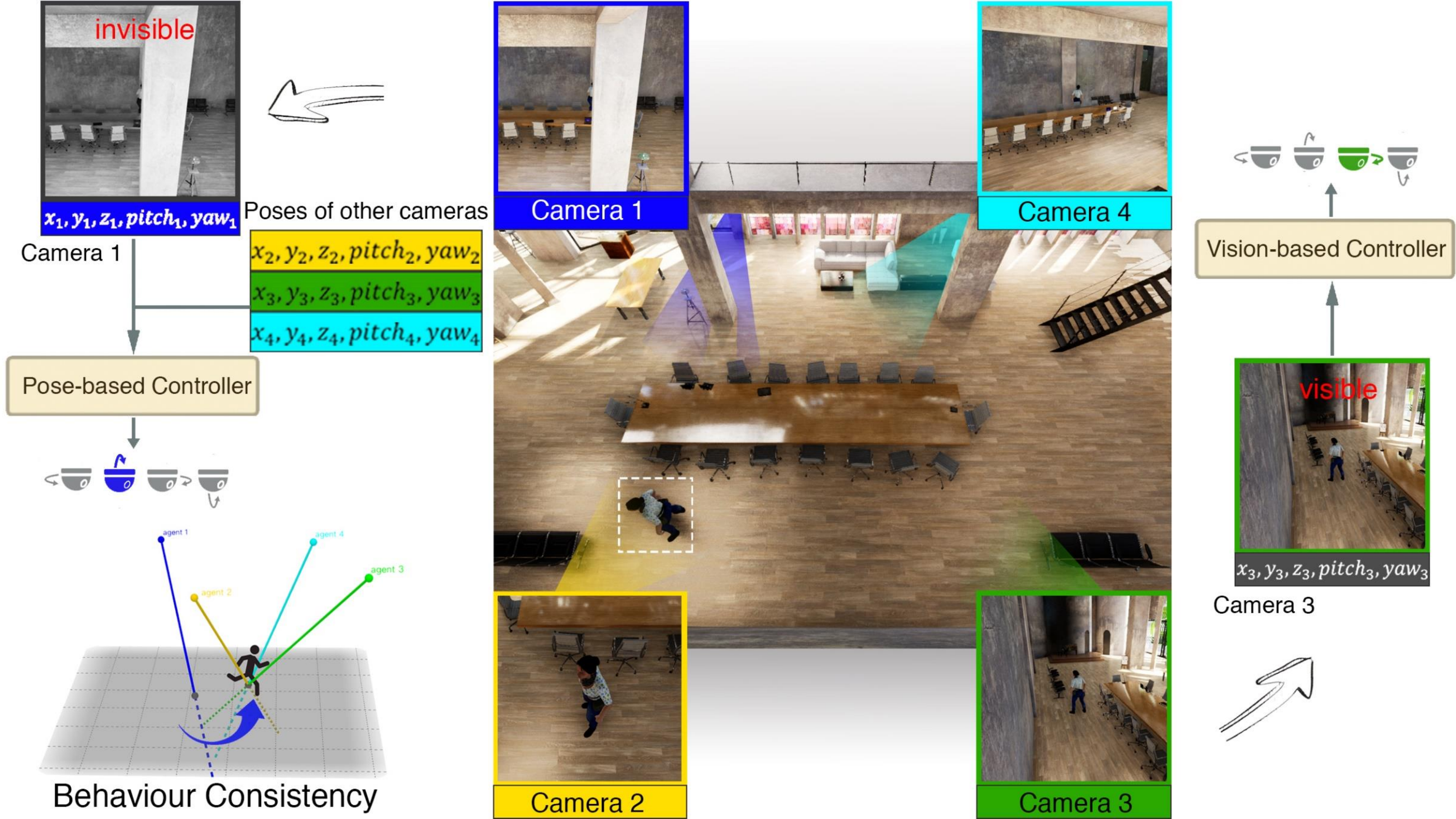}

\caption{
The overview of our multi-camera collaboration system.
When the target is visible, the camera uses its vision-based controller to make an action decision. Otherwise, the camera chooses the pose-based controller. For example, the vision-based controller of camera 1 fails to track. Thus, it uses the output action of the pose-based controller instead, and the useless visual observation is marked as gray.}
\label{fig:method}
\end{figure}

Active Object Tracking (AOT) is a fundamental and practical skill for an intelligent visual system.
It requires that a tracker be able to control its motion so as to follow a target autonomously. 
In recent years, AOT has been widely deployed in various real-world applications, such as controlling a mobile robot to follow a moving target for taking cinematic shots~\cite{hong2018virtual,luo2018end} or rotating a 3-axis stabilized camera to auto-track a face or a pedestrian. 

Applying active tracking in the surveillance scenario, rotating the camera actively, can continuously track the target in long-term.
However, there are two factors to prevent us from developing a practical AOT system for surveillance scenarios:
1) high complexity of environments, in which there are occlusion, illumination variations, scale variations of targets and obstacles, and other factors that make the observation imperfect. 
2) limitation of camera mobility, \ie the camera is only allowed to rotate and unable to shift as mobile robots do. Hence, it is particularly hard for a single-camera system to accomplish AOT in complex environments. 

We argue that it is desirable to do AOT by deploying multiple cross-view cameras and train them to work collaboratively.
Since, the trackers are able to benefit from the complimentary information provided by other cameras.
Here, we extend the single-camera AOT to a \emph{Collaborative Multi-Camera Active Object Tracking(CMC-AOT)} problem, which aims at coordinating multiple cameras in one system to improve the performance of the active tracker.
However, it is expensive and difficult to learn a general collaborative protocol under the high-dimensional visual observation, especially with the increasing number of cameras.

Therefore, in this paper, we focus on building an efficient yet effective multi-camera collaboration system for CMC-AOT.
We propose a ``Pose-Assisted Multi-Camera Collaboration System,'' which exploits the intrinsic relationship among the poses of cameras to further improve the tracking policy, shown as Fig.~\ref{fig:method}.

Inspired by the behavior of the two eyes of human beings, when tracking a target, two eyes consistently point to the same spot where the target is \--- we call the coordination of multiple cameras as \emph{Behavior Consistency}. To enable such consistency, each camera in the proposed system is equipped with two controllers (a vision-based controller and a pose-based controller) and one switcher.
The switcher selects which controller to use according to the visibility of the target in its captured image.
When the target is visible, the vision-based controller is adopted.
When the target is occluded, the switcher will switch to the pose-based controller, replacing the vision-based one. Actually the pose-based controller learns a policy about the behavior consistency, based on the poses and the switcher conditions of all cameras.
It aims at keeping the camera pose consistent with others, \ie pointing to the same area as other cameras who can observe the target.
Instead of sharing high-dimensional visual representations, our method only needs to share the poses and the conditions of switcher (indicates if the target is visible) among cameras. 

We build a set of virtual environments for training and evaluating our proposed system. The environments show high fidelity, aiming to mimic the real-world multi-camera active tracking scenarios. Specifically, we build a \emph{Random Training Room} where we randomize the surface textures, the illuminations, the sizes and locations of the obstacles, the trajectories of the target, and the distribution of cameras, \etc And we apply the A3C algorithm to update the network architecture of the Pose-Assisted Multi-Camera Collaboration System. To evaluate the generalization of the system, we build two additional realistic virtual environments, \emph{Urban City} and \emph{Garden}. Empirical results demonstrate that the learned multi-camera collaboration could generalize to unseen scenarios well and outperforms the baseline methods.
In particular, when the target is out of the view, we observe that the tracker learns to switch to the pose-based controller and keep tracking successfully.
Benefited from the multi-camera collaboration, we also find that the tracker is capable of pointing to the target even when the target is occluded.
Moreover, we conduct an ablation study to analyze the contribution of each proposed component in our system. Our method outperforms all the ablative methods in both of the testing environments, and the empirical results validate the effectiveness of the proposed method for the CMC-AOT task.

Our contributions can be summarized in three-fold:
\begin{itemize}
    \item Considering the limitation of AOT in surveillance, we extend the independent AOT to the Collaborative Multi-Camera Active Object Tracking (CMC-AOT).
    \item To achieve efficient collaboration among cameras in CMC-AOT, we propose a novel Pose-Assisted Multi-Camera Collaboration System, which enables the camera to efficiently cooperate with others by sharing camera poses.
    \item We provide a set of 3D environments for the training and evaluation of multi-camera active object tracking systems so as to facilitate future research in this direction.
\end{itemize}

\section{Related Work}
Object tracking can be complex due to some factors such as object motions, occlusions, illuminations, real-time processing requirements~\cite{yilmaz2006object}. In recent years, many video object tracking algorithms~\cite{bertinetto2016fully,choi2018context,danelljan2016beyond,held2016learning,kalal2012tracking,kiani2017learning,li2018siamrpn++,li2018high,ma2015hierarchical,valmadre2017end,wang2018learning,zhu2018distractor} have been proposed for videos, where the the motion of the camera are uncontrollable.
However, there are two typical scenarios requiring active tracking:
surveillance scene and mobile robot tracking. 
The difference between them is the action space to control the camera, \ie the mobile robot is able to move freely to choose a good perspective to track, whereas, in the surveillance scene, the camera is fixed-position and only allowed to rotate itself. 
Previous work~\cite{murray1994motion,sankaranarayanan2008objectDT,wang2013intelligent} for active tracking need two separate steps: tracking and control. The tracking step detects the target object and predicts its location. Then the control step uses the object location obtained from the track step to control the camera. But those non-end-to-end solutions require a manual bounding box to indicate the object to be tracked at the beginning. Simultaneously, joint tuning of visual tracking and camera control is also tedious and expensive. Recently, an end-to-end solution~\cite{luo2018end,luo2019end,zhong2019advat,zhong2020advat} explores under a single-target movable-camera tracking setting in which the camera is controlled by the action output from a Conv-LSTM network given raw input frames. The network is learned by reinforcement learning in a virtual environment with environment augmentation techniques.

The tracking methods above are able to track the target object based on the appearance and motion of object. But they often fail when the visual observation is imperfect \eg the object is too small to recognize, the object is occluded or the environment contains a lot of ambient noise. In contrast, our method addresses this problem by integrating pose information which acts as a complementary to image feature when an imperfect observation occurs.

Using multi-camera information to enhance the tracking performance has been investigated, \eg\cite{kang2003continuous} perform occlusion handling, accurate motion measurements, and camera hand-off through the fusion of multiple cameras. And data fusion is mainly based on the traditional triangulation approach~\cite{collins2001algorithms,kang2003continuous} which projects the location of the object in the individual image planes to the ground plane. While those traditional collaboration methods are based on image features only, thus would also fail easily with highly imperfect observations especially there are heavy occlusions. In contrast, our proposed Pose-Assisted Multi-Camera Collaboration System is to address the problem effectively by automatically switching the use of the image features and pose relationship. 
It is feasible to deal with imperfect observations and perform well in the multi-camera active object tracking problem.

\begin{figure*}[tb]
\centering
\includegraphics[width=.95\linewidth]{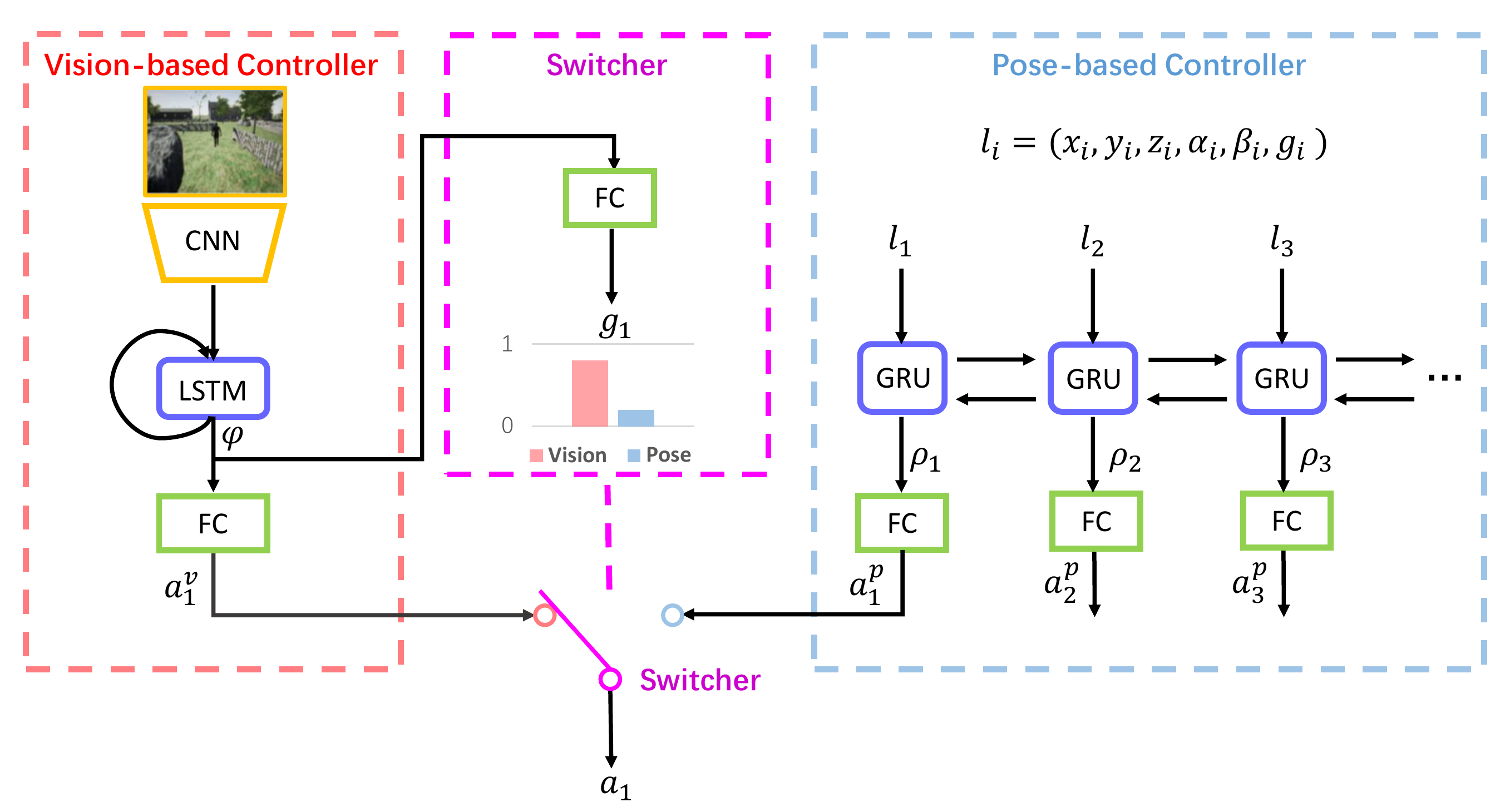}
\vspace{-0.5cm}
\caption{
The network architecture of our system. 
For each camera, the vision-based controller takes raw image as input and outputs action recommendation $a_{i}^{v}$ from its policy network, the pose-based controller takes poses and switcher labels $g$ of all cameras as input and outputs action recommendation $a_{i}^{p}$ from its policy network, then the switcher outputs the final action $a_{i} = a_{i}^{v}$ or $a_{i} = a_{i}^{p}$ by choosing to use vision-based controller or pose-based controller. Note that CNN, FC, LSTM, GRU represent Convolutional neural network, Fully connected layer, Long short-term memory, Gated recurrent unit, respectively.
}
\vspace{-0.3cm}
\label{fig:network}
\end{figure*}

\section{Methodology}
\subsection{Preliminaries}
\paragraph{Formulation.} We formulate the multi-camera active object tracking problem as a \emph{Partially Observable Multi-Agent Cooperative Game}~\cite{srinivasan2018actor} which extends the Markov Game~\cite{littman1994markov} to partial observation.
The $n$-agent game is governed by the tuple $<I, S, O_{I}, {A_{I}},{R_{I}}, T, Z>$, 
where $I, S, O, A, R, T, Z$ denote a set of agents, state space, observation space, action space, reward function, transition function, and observation function respectively.
The subscript $i \in \{1, 2,..., I\}$ denotes the index of each agent.
The subscript $t \in \{1, 2,...\}$ denotes the time step.
In the case of partial observation, 
we have the observation $o_{i,t} = o_{i,t}(s_{t})$, where $o_{i,t} \in O_{i}$, $s_t \in S$. 
It reduces to $o_{i,t} = s_t$ in case of full observation.
In the multi-agent system, we have the joint observation $\vec{o}_t = <o_{1,t},...,o_{I,t}>$ and the joint action $\vec{a_t} = <a_{1,t},...,a_{I,t}>$.
When agents take simultaneous actions $\vec{a}_t$, the updated state $s_{t+1}$ and joint observation $\vec{o}_{t+1}$ are drawn from the environment transition function $T(s_{t+1} |s_t, \vec{a}_t)$ and the observation function $Z(\vec{o}_{t+1}| s_{t+1}, \vec{a}_{t})$.
Meanwhile, each agent receives a reward $R_{i,t} = R_{i,t}(s_t, a_{i,t})$.
The policy of each agent, 
$\pi_{i}(a_{i,t}|o_{i,t})$, 
is a distribution over action $a_{i,t}$ conditioned on its observation $o_{i,t}$.
In the cooperative game, the ultimate goal is to optimize the policy of each agent to maximize the expected global reward:

\begin{equation}
    \mathbb{E}_{\pi_{1},...,\pi_{I}} \left[ \sum_{i=1}^{I}\sum_{t=1}^{T} R_{i,t} \right]
\end{equation}

Specifically, in our task, observation is the raw-pixel image and the cameras' poses, action is the camera's rotating angles in two axis (row, pitch) or zooming in/out operation.

\paragraph{Reinforcement Learning.}
We utilize the Reinforcement Learning (RL) to optimize the policy of each controller for two reasons: 1) in active tracking, the camera is a goal-directed agent interacting with the environment. The camera senses the state and takes action that affects the state. In interactive problem, it is impractical to obtain examples of all the situations, thus it needs to learn from its own experience.
2) In CMC-AOT, among the interaction with environment, there are many failure situations \ie the target disappears in the image, the target is fully occluded by occlusion in the image \etc In these situations, there is delayed reward for camera to make a right decision, thus it needs to optimize the long-term cumulative rewards to track the target in the situation with imperfect observation.

\subsection{Pose-Assisted Multi-Camera Collaboration}
In this section, we give a detailed interpretation of our Pose-Assisted Multi-Camera Collaboration System. 
The overview of our proposed network architecture is shown in Fig~\ref{fig:network}. 
In the system, each camera is equipped with three components: a vision-based controller, a pose-based controller, and a switcher. 
At each time step, the vision-based controller works independently. Meanwhile, the switcher judges if the vision-based controller is tracking the object successfully and decides to use the action of the pose-based controller or not. If the vision-based controller is considered failed, the behavior of camera will be controlled by the pose-based controller. The pose-based controller takes the outputs of the switcher and the poses of cameras as the inputs and corrects the pose of wrong camera by referring the poses of other right cameras.

\paragraph{Vision-based Controller.}
For each camera, it has a vision-based controller which serves as an image processor and guides the camera to execute policy based on image observation. The vision-based controller needs to encode the visual appearance and be aware of the motion of the target. Therefore it contains an appearance encoder $f_a(\cdot)$ and a sequence encoder $f_s(\cdot)$. The appearance encoder $f_a(o_t)$ extracts feature $\phi_t$ from the raw observation input $o_t$. Later the sequence encoder $\psi_t=f_s(\phi_1, \phi_2, ..., \phi_t)$ models the temporal differences over time. In this way, it encodes a representation $\psi_t$ containing temporal differences information which is fed into the subsequent policy network, and the policy network outputs camera's action $a_{i}^{v}$ at each time step.

Concretely, Convolutional Neural Networks (CNNs) are used here by the appearance encoder $f_a$. And the sequence encoder is a LSTM Network to deal with sequential features $(\phi_{i,1},\phi_{i,2}, ..., \phi_{i, t})$ and outputs the final hidden features to the policy network.

\paragraph{Pose-based Controller.}
The goal of the pose-based controller is to help the camera who receives an imperfect observation to execute policy based on the supplementary pose information provided by other cameras. 

There is \emph{Behavior Consistency} among AOT cameras, \ie they all need to point to the same area where the target is. Inspired by this, the pose-based controller is constructed to exploit the behavior consistency of cameras, and then guides the camera with imperfect observation to rotate to the right pose.
We equip each camera with a binary label \ie $g_{i} \in \{0, 1\} $, provided by the switcher which indicates if the camera is tracking the object successfully by the vision-based controller at the current time step. The pose-based controller takes all binary labels and poses of the camera as input, containing the successful cameras and failed cameras, aiming to guide the failed camera to rotate in the right direction under hints given by the poses of other successfully tracking cameras.

Specifically, the camera pose we use here contains location information of camera and its rotation $ l_{i} = \{x_{i}, y_{i}, z_{i}, \alpha_{i}, \beta_{i}\}$ where $x_{i}, y_{i}, z_{i}$ denote camera i's location, $\alpha_{i}$ denotes the its pitch angle, $\beta_{i}$ denotes its yaw angle. The pose encoder $f_{p}$ for which we use a bidirectional Gated Recurrent Unit (GRU) network encodes the poses and binary labels of all cameras into pose features $ \rho_{1},\dots, \rho_{I} = f_{p}(l_{1}, g_{1}, \dots, l_{I}, g_{I})$. For each camera, its received pose feature from the pose encoder contains instructive information of other cameras which can help it execute right policy when its vision-based controller failed. 

Introducing the camera pose into the CMC-AOT problem is beneficial and it has two advantages: \emph{instructive} and \emph{low-cost}. In terms of effectiveness, the pose information is instructive. Every camera can get indicative information from the poses of other cameras when it comes to collaborative object tracking. 
On the other hand, due to the simplicity of pose representation, processing the pose information instead of image information of the camera can save the expensive transmission cost among multiple cameras.
\vspace{-0.75em}
\paragraph{Switcher.}
Since there is no god's perspective to tell the camera when the vision-based controller is failed, we need a switcher to make the camera switch between the vision-based controller and pose-based controller properly. That is, when the vision-based controller fails, the switcher chooses to replace the vision-based controller with the pose-based controller. A well-worked switcher can switch between the camera's two controllers reasonably and properly.
Specifically, the switcher is a binary classified neural network that receives the image features from vision-based controller $\psi_{i, t}$ and outputs the probabilities of choosing controllers.  

In the CMC-AOT problem, our Pose-Assisted Multi-Camera Collaboration System can achieve well-coordinated collaboration under good cooperation of the vision-based controller, the pose-based controller, and the switcher.

\subsection{Reward Structure}
To learn active tracking and the multi-camera coordination simultaneously, it is essential to design an appropriate reward function for CMC-AOT.

To achieve active tracking successfully, the goal is to minimize the error between the current and expected camera pose. The expectation camera pose corresponds to the pose when the target object is located in the center of the view of camera.
Therefore, we use pose error to measure the tracking effect when the target is in the image. Note that when the target is occluded by the obstacles causing that it did not appear in the image, we set the tracking reward as $0$, while when the target goes beyond the view angle of the camera, we set the tracking reward as $-1$. We use a vector $(\alpha, \beta, \lambda)$ to represent the rotation from the frame of the camera to the world frame under the form of Euler Angles. In other words, $\alpha$, $\beta$, $\lambda$ correspond to the pitch, yaw, and roll angles. Note that for direction reward, we do not take the roll angles $\lambda$ into account, since we do not control the camera to rotate along this axis at all, we can only control pitch, yaw angles of camera in the CMC-AOT task. When the target appears in the image, we use the yaw angle and pitch angle between camera direction and target direction to calculate direction reward. The goal of optimization is to minimize the angle error as much as possible.
Concretely, the direction tracking reward is written as below:

\begin{equation}
R_{d, t} = \left\{
\begin{aligned}
 &1 - \frac{\Delta \alpha_{t}}{\alpha_{max}} - \frac{ \Delta \beta_{t}}{\beta_{max}}, & (a) \\
& 0, &  (b) \\
& -1, & (c)
\label{eq:tracking_reward}
\end{aligned}
\right.
\end{equation}

(a): target is visible in the image; (b): target is occluded by obstacles; (c): target is outside of the view.
Where $\Delta \alpha_{t}$ and $\Delta \beta_{t}$ are absolute pitch angle error and absolute yaw angle error between camera direction and target direction respectively. $\alpha_{max}$ and $\beta_{max}$ give maximum control bound of each angle error. 
Note that to achieve better tracking performance, we add zoom actions in addition, \ie choosing zoom in or zoom out action to enlarge or shrink the image, changing its zoom scale $\xi$ which ranges from 1 to 3.3.
Similarly, we compute additional zoom reward as $R_{z, t} = 1 - \frac{\Delta \xi_{t}}{\xi_{max}}$ in condition (a), otherwise $R_{z, t} = 0$, where $\Delta \xi_{t}$ is the absolute zoom scale error at time step $t$.
Hence, the reward of the camera at each step is 
\begin{equation}
    R_{t} = R_{d, t} + R_{z, t}
\end{equation}

Note that the rewards above are clipped in the range of $[-1, 1]$ while training. Under the designed reward, the more accurately the camera keeps tracking the target, the higher the cumulative reward is. 

\subsection{Training Strategy}
We take a two-phase training strategy for learning.
In the first phase, we train the pose-based controller in a numerical simulator, which only performs numerical calculations on the poses of target and cameras without rendering the image.
The pose-based controller does not need the image as input and only observe the pose (location and rotation) of each camera and the binary label of the switcher (choosing vision or pose) to rotate the camera.
Since there is no vision-based controller at this stage, we randomly set the label of the switcher at each step.
During training, if the switcher selects the pose-based controller, the camera will take the action proposed by the pose-based controller. After that, the returned reward will be used to optimize the policy network via reinforcement learning.
If not, the camera will be controlled by a ``virtual tracker", which takes the nearly optimal action to minimize the errors of relative angles between the camera and target.
To further improve the generalization, we randomize the distribution of the cameras and the trajectories of the target at each episode.
Thanks to the high frame rate of the simulator, the time cost of the training process of the pose-based controller is significantly reduced to 10 $\sim$ 15 minutes.

In the second phase, we train the vision-based controller and the switcher simultaneously, combining with the pose-based controller trained at the first phase. 
Specifically, the vision-based controller is trained by reinforcement learning, meanwhile, the switcher is regarded as an auxiliary classifier \ie predicting if the target exists in the image. 
Intuitively, the visibility of the target is closely related to the choice of the controller, \ie we use the vision-based controller when the target is visible, otherwise, we use the pose-based controller.
The switcher is optimized by binary cross-entropy loss, and the virtual environment could provide the ground-truth for learning by extracting the object mask.
Note that we backward propagate the gradients of the vision-based controller only when the switcher selects the vision-based controller, and the network parameters of the pose-based controller is frozen at this phase.

\section{Experiments}
\subsection{Environments}

\begin{figure}[tb]
\centering
\includegraphics[width=\linewidth]{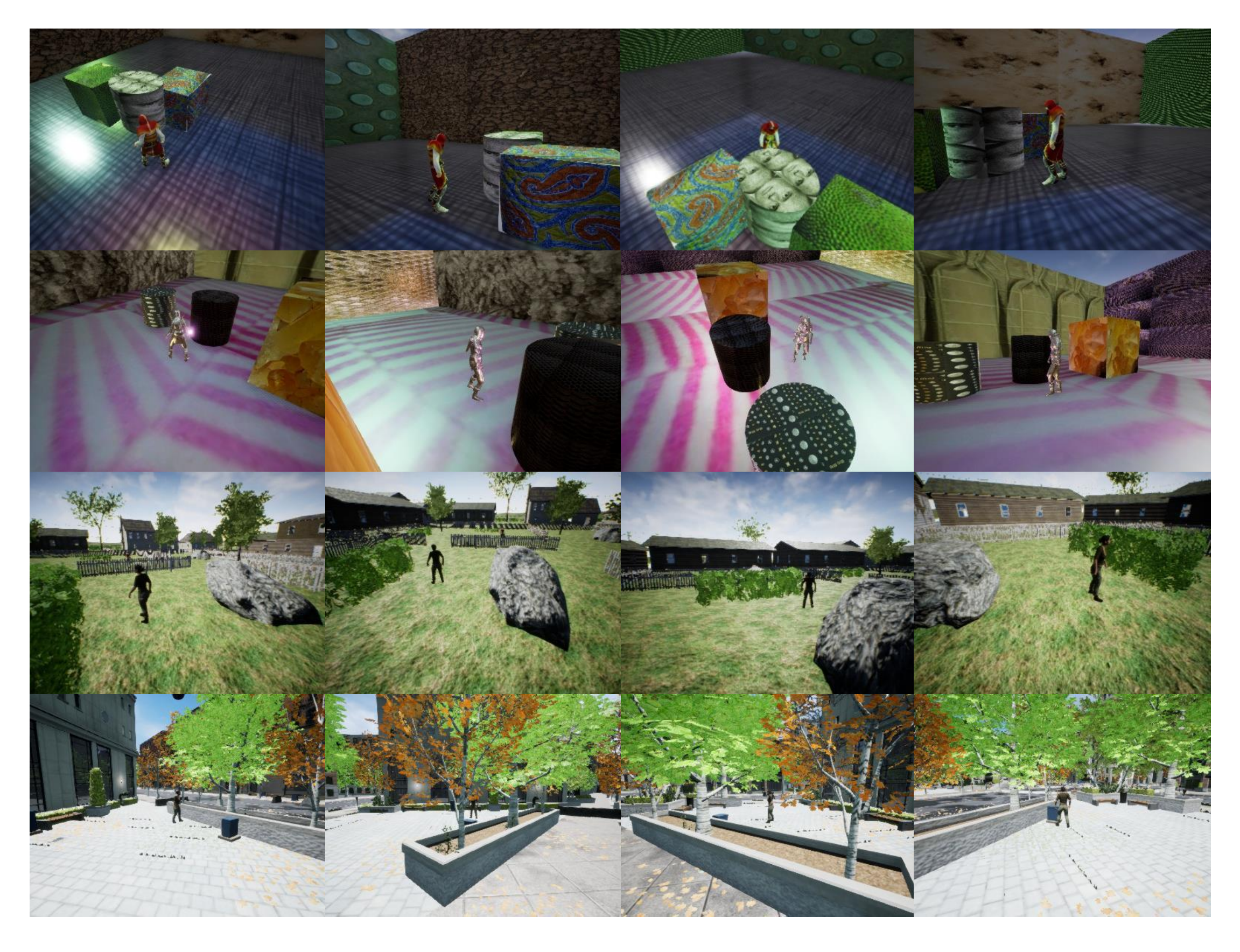}
\vspace{-0.7cm}
\caption{
From Top to bottom is the 3D environment \emph{Random Room} for training, \emph{Garden}, \emph{Urban City} and for testing. Note that our model is only trained on the \emph{Random Room}.
}
\vspace{-0.3cm}
\label{fig:env}
\end{figure}

First of all, we build a number of high-fidelity virtual environments for learning and testing.
We do this, instead of running on the real-world environment directly, for three reasons: 1) Reinforcement learning needs to interact with environment frequently and learns from trail-and-error which is high-cost in real environment; 2) In real-world scenario, it is difficult and expensive to get the ground truth to compute the reward function for training and evaluation; 3) Previous works~\cite{hong2018virtual,luo2019end,zhong2020advat} have justified that the tracker trained in virtual environment with environment augmentation is capable of generalizing to real-world scenes.
We build a number of new 3D environments for the CMC-AOT task, in which there are more cameras and more obstacles in the environment, aiming to mimic the real-world multi-camera active tracking scenes. 
The  action space is discrete and contains eleven candidate actions (\emph{turn left, turn right, turn up, turn down, turn top-left, turn top-right, turn bottom-left, turn bottom-right, zoom in, zoom out and keep still}).

For training, we build a large room with four cameras and one moving person. The four cameras are randomly placed around the room at the beginning of each episode. The person is walking in the room with random velocity and trajectories. 
To simulate the occlusion cases, we randomly place a number of obstacles with different shapes and sizes. 
To learn better feature representation in terms of visual observation, we also randomize the illumination condition and the surface textures of each object. Specifically, we choose pictures from a texture dataset~\cite{Kylberg2011c} and place them on the surface of walls, floor, obstacles \etc 
These randomization methods are referred as environment augmentation and the room with environment augmentation is named \emph{Random Room}, as shown in the top two lines of Fig.~\ref{fig:env}. 

To demonstrate the capability of transferring to unseen scenes, we also build two realistic environments -- \emph{Garden}, \emph{Urban City} as testing environments.
Both environments mimic real-world scenarios and cover the most of challenging cases in active object tracking: obstacle occlusion, illumination variation, scale variation. \emph{Garden}, as shown in the third line of Fig.~\ref{fig:env}, is grassland with trees, fences and big stones \etc
These objects frequently block the target, causing the tracker to fail to observe the target. 
The difficulty increases with the complex illumination which makes the appearance of the scene vary in a wide range across different perspectives. 
\emph{Urban City}, as shown in the last second line of Fig.~\ref{fig:env}, is a high-fidelity street view of an urban city, including well-modeled buildings, streets, trees and transportation facilities.

\subsection{Evaluation Metric}
We use the angle error between camera direction and target direction to evaluate the quality of active object tracking. The angle error is the average of absolute pitch angle error and absolute yaw angle error. A well-worked camera should track the target accurately in both the pitch angle and yaw angle. 

The multi-camera system evaluation metric is the average \emph{Single Camera Error} of all I cameras:
\vspace{-0.1cm}
\begin{equation}
    \emph{Mean Error} = \frac{1}{I}\sum_{i=1}^{I}\frac{1}{T} \sum_{t=1}^{T} 
\frac{(\Delta \alpha_{i, t}+\Delta \beta_{i, t})}{2}
\label{metric:error}
\end{equation}

where $I$ is the numbers of camera, the $\Delta \alpha_{i, t}$ and $\Delta \beta_{i, t}$ are the absolute error of pitch angle and yaw angle of camera $i$ at time step $t$, respectively. 
We use \emph{Mean Error} to measure the performance of the multi-camera system finally.

\emph{Mean Error} reflects the tracking precision of the multi-camera system. However, it is also beneficial to evaluate the multi-camera tracking success rate (\ie if there are more cameras recover tracking under collaboration). Therefore, we introduce \emph{Success Rate} to better evaluate the robustness.
\vspace{-0.1cm}
\begin{equation*}
     \emph{Success Rate} = \frac{1}{I}\sum_{i=1}^{I}\frac{1}{T}\sum_{t=1}^{T} D_{i, t}  
\label{metric:rate}
\end{equation*}

where $D_{i,t}$ equals to $1$ if the target is within the perspective of the $i$'th camera at time step $t$, otherwise equals to $0$, $T$ is the fixed episode length in the evaluation.

\begin{table*}[tb]
\centering
\caption{Comparative results (TLD, BACF, DaSiam vs Ours) and ablative results (SV, MV, SV+P vs Ours) in the 3D \emph{Garden} and \emph{Urban City} environments.}
\vspace{-0.2cm}
\label{table-garden}
\begin{center}
\setlength{\tabcolsep}{2pt}
\resizebox{\textwidth}{!}{
\begin{tabular}{|c c |c c c c c c c|c c c c c c c|}
\hline
\multicolumn{2}{|c|}{} &\multicolumn{7}{c|}{\bf Mean Error (In degrees.)}
&\multicolumn{7}{c|}{\bf Success Rate(\%)}\\ 
\hline
Env & Cam id & TLD & BACF & DaSiam & SV & MV  & SV+P & Ours  & TLD & BACF & DaSiam & SV & MV  & SV+P & Ours \\
\hline
\hline
\multirow{4}{*}{Garden}
 &Cam\_1      &28.53     & 32.74  & 25.73 &  29.07  &  36.30 & 31.53 & \bf 10.95 &58.39  & 50.62  & 67.81 & 66.67 & 42.62 & 54.34 & \bf 83.94 \\
 &Cam\_2      &21.90     & 21.19  & 20.42 &  26.42 &  28.15 & 26.32 &  \bf 9.81 &73.26  & 70.91  & 75.23 & 64.18 & 46.37 & 65.79  & \bf 88.60 \\
 &Cam\_3      &33.46     & 34.42  & 32.15 &  40.15 &  30.65 & 35.19 & \bf 13.64 &53.69  & 45.47  & 54.79 & 46.67 & 44.95 & 42.94 & \bf 80.44\\
 &Cam\_4      &21.47     & 22.17  & 17.86  &  16.65 & 32.14 & 14.16 & \bf 8.62 &73.74   & 63.92  & 78.41 & 80.76 & 39.90 & 88.97 & \bf 89.65 \\
\hline 
\multicolumn{2}{|c|}{Mean} &26.34     & 27.63  & 24.04 &  28.07  &  31.81 & 26.80 & \bf 10.76 &64.77    & 57.73  &  69.06  & 64.57 & 43.46 & 63.01  & \bf 85.66  \\
\hline
\multirow{4}{*}{Urban City}
&Cam\_1      &23.68     & 28.75  & 21.34 & 26.18    & 16.02 & 21.73 & \bf 7.22 &68.83     & 59.23  & 73.05 & 58.05 & 72.59 & \bf 78.24 & 75.19  \\
&Cam\_2      &21.43     & 27.90  & 19.78 & 13.11 & 20.16 & 20.04 &\bf  8.10 &73.42     & 59.01  & 75.32  & 77.23 & \bf 86.69 & 79.35  & 82.78\\
&Cam\_3      &54.39     & 66.41  & 49.85 & 62.46    & 30.36  & 38.60 & \bf10.02  &10.13   & 6.16  & 12.93 & 29.10 & 31.65 & 45.62 & \bf89.33\\ 
&Cam\_4      &25.18     & 32.98  & 20.75 & 34.31  & 21.29 & 21.87 & \bf6.74 &69.58     & 49.64  & 74.66 & 43.58 & 63.17 & 72.71  & \bf89.41\\ 
\hline 
\multicolumn{2}{|c|}{Mean} &31.17  & 39.01  & 27.93 &  34.02 &  21.96 & 25.56  & \bf 8.02 &55.49  & 43.51 & 58.99 & 51.99 & 63.53 & 68.98 & \bf 84.18  \\
\hline
\hline
\multicolumn{2}{|c|}{Average} &  28.76   & 33.32  & 25.99 &  31.05  &  26.89 & 26.18  & \bf 9.39 &60.13     & 50.62  & 64.03 & 58.28 & 53.50 & 66.0 & \bf84.92  \\
\hline
\end{tabular}
\label{tab:all_results}
}
\vspace{-0.3cm}
\end{center}
\end{table*}

\subsection{Compare with Two-stage Methods}
We compare our method with conventional two-stage tracking methods, \ie the controller rotates the camera according to the bounding box of the target from a video tracker.
We adopt three video tracker to get the bounding box: TLD~\cite{kalal2012tracking}, BACF~\cite{kiani2017learning}, and DaSiamRPN~\cite{zhu2018distractor}. 
We build a heuristic controller to control the camera to rotate its angle.
The control policy is based on a rule that the camera moves its angle along with the position of the detected bounding box, \ie when the bounding box is at the left of the image, the controller output a turn-left signal.

Table~\ref{tab:all_results} shows the concrete results of \emph{Mean Error} and \emph{Success Rate} evaluation metrics on \emph{Garden} and \emph{Urban City} environments. We can see that the traditional tracking methods perform weakly in the CMC-AOT system. 
We analyze that the traditional tracker suffers from two problems that usually make it fail: 1) the object appearance changes largely and 2) the target is frequently occluded by obstacles. Since these methods require the template for object feature matching, when the object disappears outside the view of the camera, the tracker has no effective information on the image to use which leads to tracking mistakes easily. Once the tracker lost the target, the tracking error will increase largely in the long-term. While in our method, we learn the sequential feature information by the vision-based controller which makes our tracker becomes more robust with various environment. More importantly, our method can handle scenes with obstacles or other imperfect observations well due to the assistance of the pose-based controller. And the experiment results demonstrate that our method is significantly superior to traditional tracking methods.

\begin{figure*}[tb]
\centering
\includegraphics[width=0.24\linewidth]{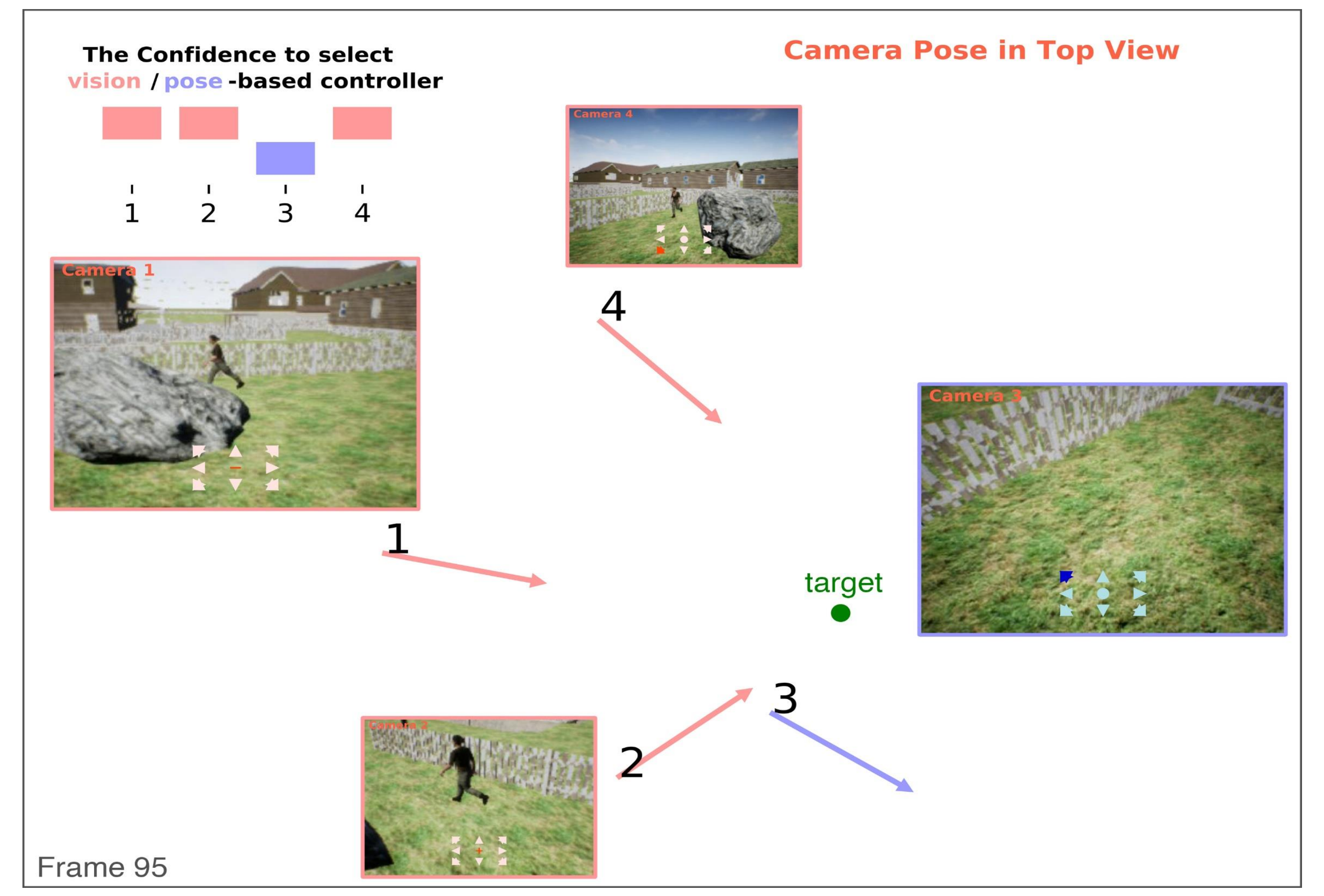} 
\includegraphics[width=0.24\linewidth]{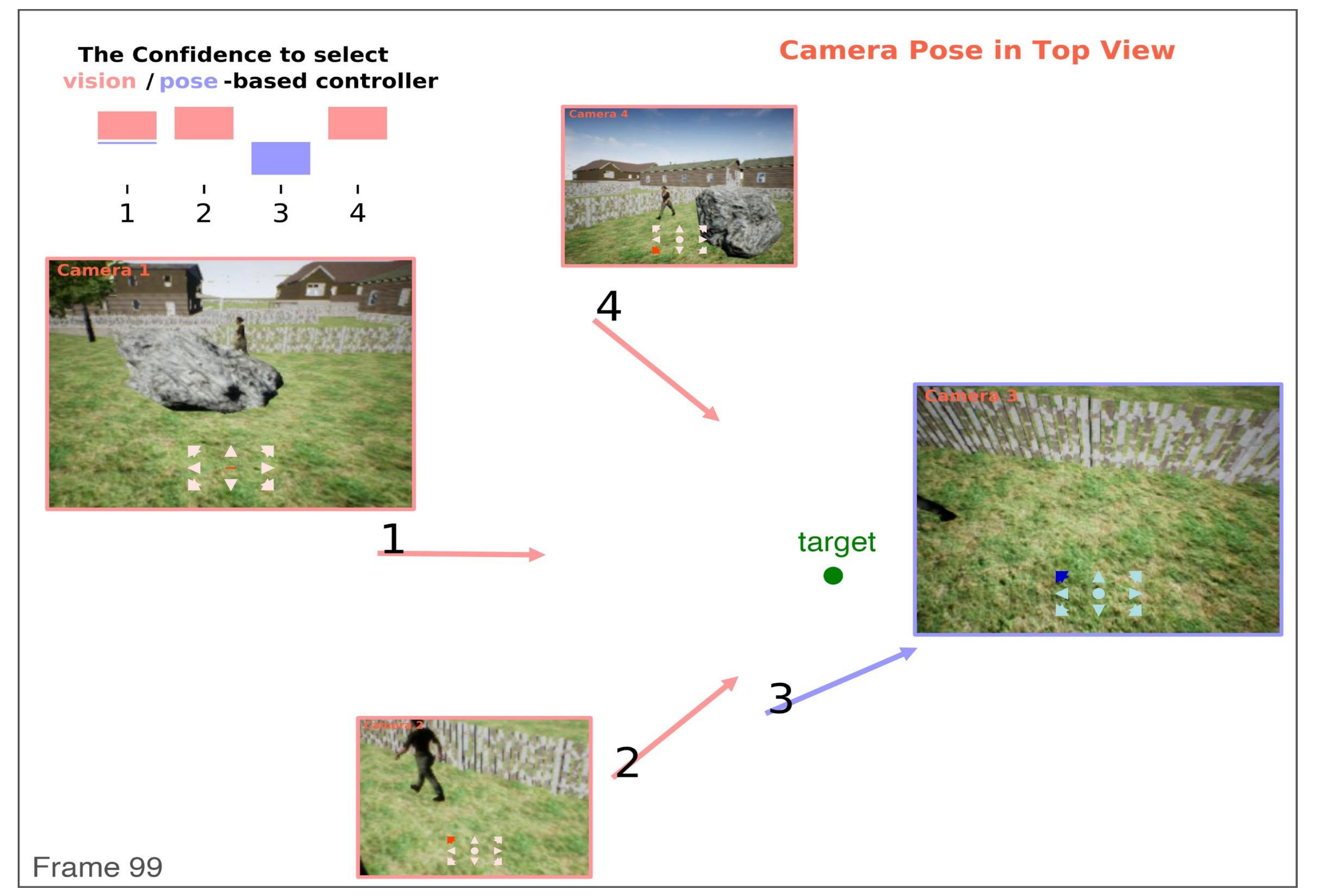}
\includegraphics[width=0.24\linewidth]{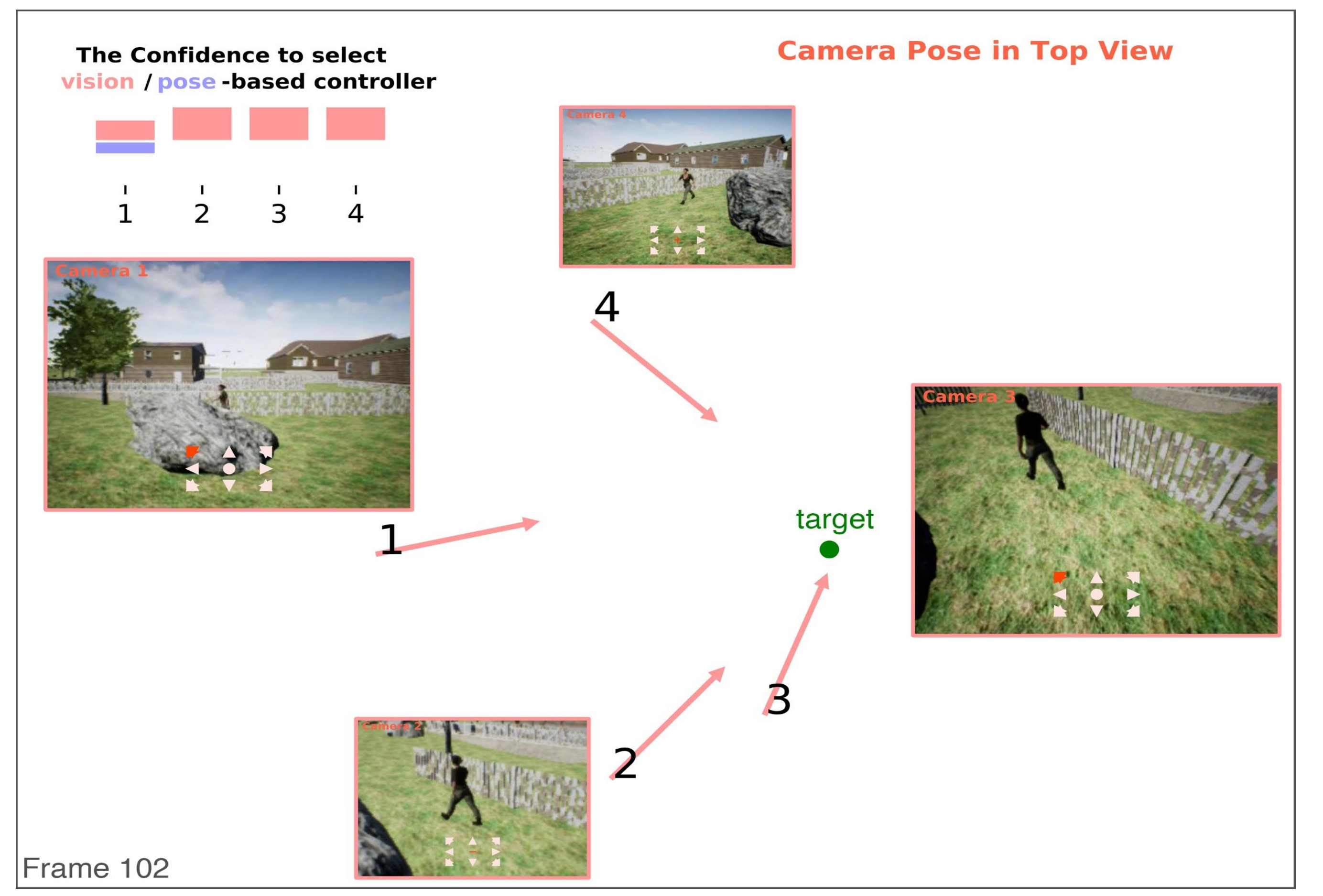}
\includegraphics[width=0.24\linewidth]{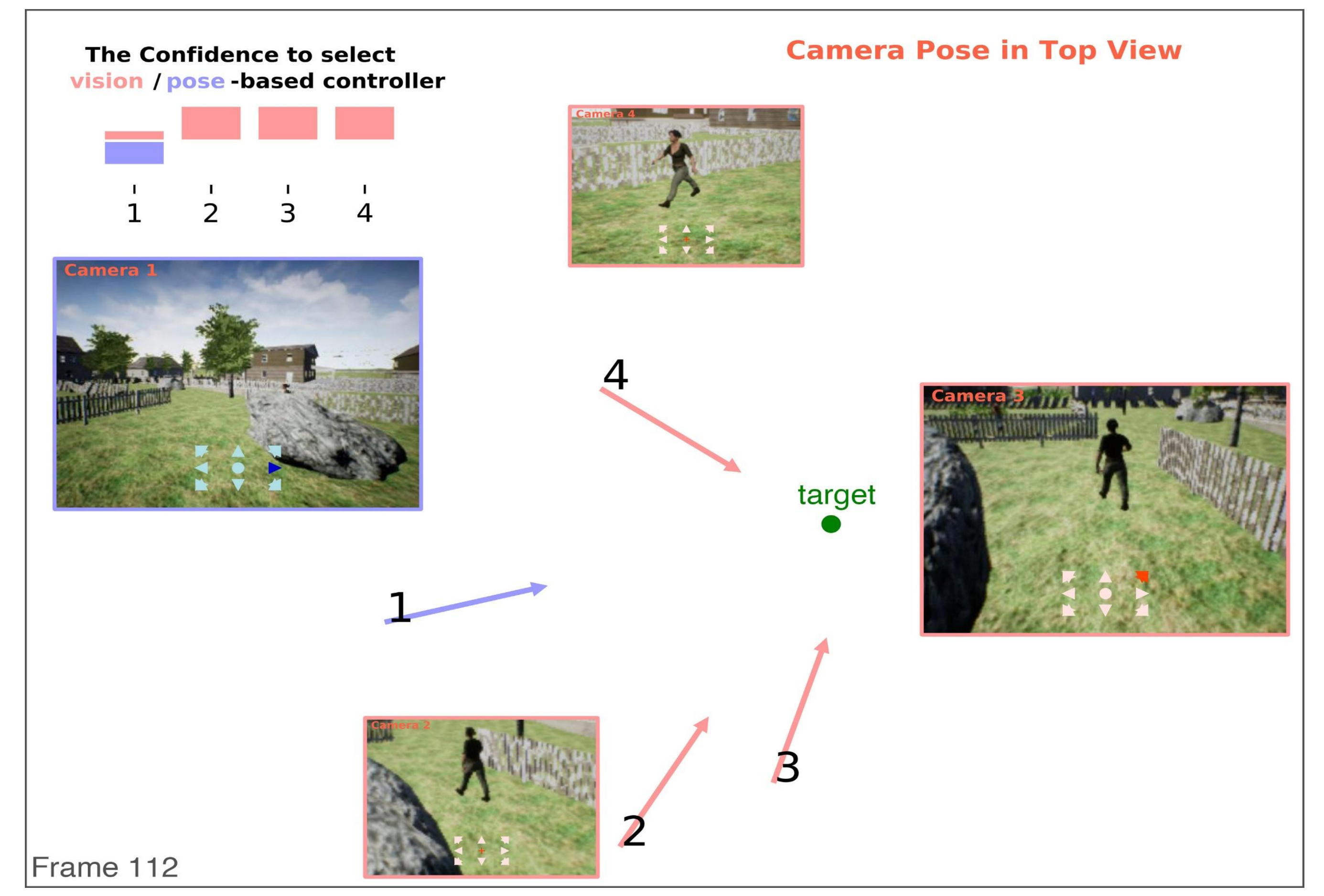}
\vspace{-0.3cm}
\caption{The screenshot sequence of our method working in the \emph{Garden} testing environment. The arrow position and direction indicate the location and the pointing direction of the camera. The four frames are visual observations of four cameras respectively. The eight direction signs indicate rotating direction of action, the intermediate sign indicates keep still action when it is a circle, zooming out when it is a minus sign and zooming in when it is a plus sign. Red means using output action of the vision-based controller, blue means using output action of the pose-based controller.}
\label{fig:garden_seq}
\vspace{-0.4cm}
\end{figure*}

\subsection{Ablative Analysis}
To analysis the effectiveness of the proposed Pose-Assisted Multi-Camera Collaboration System quantitatively, we conduct ablative experiments with different collaboration methods. Concretely, we use the ablative analysis aiming to answer three concerns: 1) is the multi-view collaboration necessary? 2) is the introduced pose information useful for CMC-AOT? 3) is our method effective to exploit the pose for collaboration?
For this purpose, we further introduce and evaluate three heuristic neural architectures:
\begin{itemize}
    \item Single View (SV): Each tracker only uses its own visual observation to control the camera independently. The neural network is the same as the end-to-end vision-based controller in Fig.~\ref{fig:network} which is first proposed in \cite{luo2018end}.
    \item Multiple View (MV): Each tracker fuses the visual representation from others as the input of the policy network. Similar to the pose encoder, the multi-view representation is also integrated by a one-layer Bi-GRU. 
    \item Single View with Poses (SV + P): Each tracker encodes a representation which aggregates its visual observation and the poses of other cameras by Fully Connected Layer to its policy network. The relative poses are encoded by a one-layer Bi-GRU.
\end{itemize}

Comparing SV with MV, the results show that fusing multi-view features directly is not effective. SV uses only the vision-based controller at each step without any information transmission between agents.
The multi-view fusion method implements the image feature information exchange between cameras by direct network fusion. However, sharing the visual information in this way is time-consuming and not efficient. There is no obvious improvement, showing that this collaboration way can not work well in the CMC-AOT system.  
Comparing SV and SV + P, the results show that introducing
pose into the collaboration by directly fusing the vision feature and pose feature can not improve performance obviously. Fusing pose information to the collaboration method is still indistinct. While comparing SV + P with our method, it shows that introducing pose into our multi-camera tracking system by a switcher can get a significant performance of collaboration tracking.
We use such a designed collaboration structure to explicitly and efficiently ensemble two controller’s policy. The clear and effectiveness of each module makes our collaboration method not limited by the imperfect situations and can achieve higher tracking accuracy significantly. The lower \emph{Mean Error} and higher \emph{Success Rate} in Table~\ref{tab:all_results} shows the higher tracking accuracy of our method.
\vspace{-0.3cm}
\paragraph{Exemplar Cases.}
We take a sequence shown in Fig~\ref{fig:garden_seq} to demonstrate how our three modules work cooperatively. There exist many obstacles such as brushwood, big stones, railing etc in the garden scene. It is difficult to track the target person accurately since the obstacle will occlude the person easily, as shown in camera 1 of frame 112, the person walks behind the big stone. The vision-based controller of the camera fails to track the target, and its switcher makes it turn to the pose-based controller at the time, hence the camera can keep tracking the target people successfully even the person is occluded by a big stone.
Our method can work not only in occlusion situations but also in other imperfect observation situations. As shown in frame 95, the vision-based controller of camera 3 lost the target. In this situation, it is difficult for the tracker to make action decisions only based on the imperfect image observation. Thus the switcher of camera 3 chooses to use the pose-based controller which helps it recover the person successfully, as shown in frame 99 and frame 102. In a word, by the coordination of the vision-based controller, the pose-based controller and the switcher in the system, our collaboration approach combines the advantages of the image and pose of camera which can improve the overall performance in the CMC-AOT system greatly. To see more cases, please refer to the demo video on the homepage of our project.

\vspace{-0.2cm}
\section{Conclusion}
In this work, we introduce the Collaborative Multi-Camera Active Object Tracking (CMC-AOT) problem, and propose an effective Pose-Assisted Multi-Camera Collaboration System to further enhance the tracking performance. By introducing camera pose into the multi-camera collaboration, our method has the ability to deal with challenging scenes and outperforms traditional object tracking methods on a variety of multi-camera active object tracking environments. The results on different realistic environments also show that our approach has the potential to generalize to more unseen scenes. 

\vspace{-0.2cm}
\section{Acknowledgments}
We thank Tingyun Yan and Weichao Qiu for providing assistance with UnrealCV.
This work was supported by MOST-2018AAA0102004, NSFC-61625201, NSFC-61527804, Qualcomm University Research Grant.

\begin{quote}
\begin{small}
\bibliographystyle{aaai}
\bibliography{aaai}
\end{small}
\end{quote}

\end{document}